# Understanding Clinical Decision-Making in Traditional East Asian Medicine through Dimensionality Reduction: An Empirical Investigation

Hyojin Bae[a,†], Bongsu Kang[b,†], and Chang-Eop Kim[b*]

[a] Department of Physiology, Seoul National University College of Medicine, 03080 Seoul, Korea

[b] Department of Physiology, Gachon University College of Korean Medicine, 13110 Seongnam, Korea

†These authors contributed equally to this work.

* Corresponding author: Chang-Eop Kim, Email: eopchang@gachon.ac.kr

## Abstract

This study examines the clinical decision-making processes in Traditional East Asian Medicine (TEAM) by reinterpreting pattern identification (PI) through the lens of dimensionality reduction. Focusing on the Eight Principle Pattern Identification (EPPI) system and utilizing empirical data from the *Shang-Han-Lun*, we explore the necessity and significance of prioritizing the Exterior-Interior pattern in diagnosis and treatment selection. We test three hypotheses: whether the Ext-Int pattern contains the most information about patient symptoms, represents the most abstract and generalizable symptom information, and facilitates the selection of appropriate herbal prescriptions. Employing quantitative measures such as the abstraction index, cross-conditional generalization performance, and decision tree regression, our results demonstrate that the Exterior-Interior pattern represents the most abstract and generalizable symptom information, contributing to the efficient mapping between symptom and herbal prescription spaces. This research provides an objective framework for understanding the cognitive processes underlying TEAM, bridging traditional medical practices with modern computational approaches. The findings offer insights into the development of AI-driven diagnostic tools in TEAM and conventional medicine, with the potential to advance clinical practice, education, and research.

## 1. Introduction

Traditional East Asian Medicine (TEAM) has been practiced for millennia, offering a unique and structured approach to diagnosing and treating illnesses(1). A key aspect of clinical decision-making in TEAM is pattern identification (PI). This process classifies a patient's symptoms into predefined groups, known as patterns or syndromes, to form a coherent diagnosis and treatment plan(2). In a previous study, we introduced a novel framework that reinterprets PI through the lens of machine learning, proposing that the PI process can be understood as a form of dimensionality reduction(3). In this context, PI is conceptualized as the process of transforming complex, high-dimensional symptom data into a more manageable, low-dimensional representation, thereby revealing latent patterns that may not be immediately apparent. This approach suggests that the effectiveness of PI in TEAM may stem from its ability to simplify and distill critical information from a vast array of symptoms, ultimately reducing cognitive load for TEAM doctors and aiding in their decision-making process. This perspective provides a foundation for a quantitative and systematic understanding of how diagnostic processes in TEAM are derived from clinical observations.

Building upon this foundation, the present study aims to empirically examine the unique information-processing characteristics inherent in PI through quantitative analysis techniques. To explore this in detail, we utilized the authoritative clinical text in TEAM, *Shang-Han-Lun*, and the Eight Principle Pattern Identification (EPPI) system, one of the most comprehensive PI systems in TEAM(4, 5). EPPI represents a patient's symptoms as a combination of three patterns, each indicating a specific pathological state based on a dichotomous interpretation of the Yin-Yang principle. This approach allows EPPI to project the wide range of symptoms exhibited by patients onto three key dimensions: Exterior-Interior, Cold-Heat, and Deficiency-Excess. In this dimensionality reduction process, it is crucial to note that TEAM practitioners are guided to first evaluate the depth of disease or pathogenic factors in the human body (Exterior-Interior pattern) before assessing the other two patterns(6). This raises important research questions: Is the precedence of the Exterior-Interior pattern in the diagnostic process justified? And is this prioritization essential for an effective clinical decision-making process?

To address these questions, we proposed three hypotheses regarding the role of sequence in the PI process and explored each in detail. First, we hypothesized that the Exterior-Interior dimension would contain the most information about the patient's symptoms, based on the assumption that it would capture the greatest variance in the symptom data. Second, we anticipated that the Exterior-Interior dimension would represent the most abstract information about the symptoms, as highly abstract representations facilitate efficient classification and generalization to new cases. Furthermore, we hypothesized that the Exterior-Interior dimension would also encapsulate the most abstract information about herbal prescriptions as well, thereby enhancing the accuracy of treatment selection. This hypothesis is grounded in the core function of PI: mapping a patient's symptoms to an appropriate treatment.

This study demonstrates that the information processing and clinical decision-making in TEAM, from diagnosis to treatment, can be effectively implemented and understood using quantitative methodologies. Measures including the abstraction index, cross-conditional generalization performance, and decision tree regression were employed to test our hypotheses and provide an explainable, objective theory. Furthermore, by mathematically modeling the cognitive processes of TEAM practitioners, this paper provides insights into how TEAM doctors process information to make clinical decisions and what benefits arise from using PI. This research contributes to a deeper understanding of PI and TEAM, and it helps bridge the gap between traditional medical methodologies and modern, objective interpretations.

## 2. Results

### 2.1. Hypotheses

#### 2.1.1. Hypothesis I: The Exterior-Interior dimension contains the most information about the patient's symptoms.

Among the Eight Principle Patterns of EPPI, six are composed of the following: Exterior (Ext, 表), Interior (Int, 裏), Cold (寒), Heat (熱), Deficiency (Def, 虛), and Excess (Exc, 實). The remaining two, yin (陰) and yang (陽), are overarching concepts that encompass all six of these patterns(7). The six principle patterns are organized into three pairs of mutually opposing properties, which serve as dimensions in a reduced space: Ext-Int, Cold-Heat, and Def-Exc. These three pairs correspond to the disease's location, the nature of the disease, and the relative dominance between pathogenic factors and anti-pathogenic qi (氣), respectively.

A common and effective way to estimate the amount of information in a dataset is to assess the variance of data along different axes. The greater the variance along an axis, the more information that axis is likely to capture(8). Therefore, the axis that accounts for the highest proportion of variance in the data space can be considered the most informative dimension.

As depicted in **Fig. 1A**, the data manifold representing the spanned space in the pattern space is not a complete sphere, indicating that the subspaces are not uniformly spanned. Specifically, the subspaces composed of Int-Exc, Int-Heat, and Def-Cold are densely spanned, while those composed of Ext-Exc, Ext-Cold, and Def-Heat are more sparsely spanned. Among the combinations of dimensions, a positive correlation between the Cold-Heat and Def-Exc dimensions is most prominent (**Fig. 1B**). Interestingly, the univariate distribution for the Ext-Int dimension had the smallest variance and was the most skewed, which was contrary to our expectation (**Table 1**). This finding suggests that hypothesis I is not supported, as the Ext-Int dimension, with its smallest variance, appears to capture the least amount of information about the symptoms compared to the other dimensions.

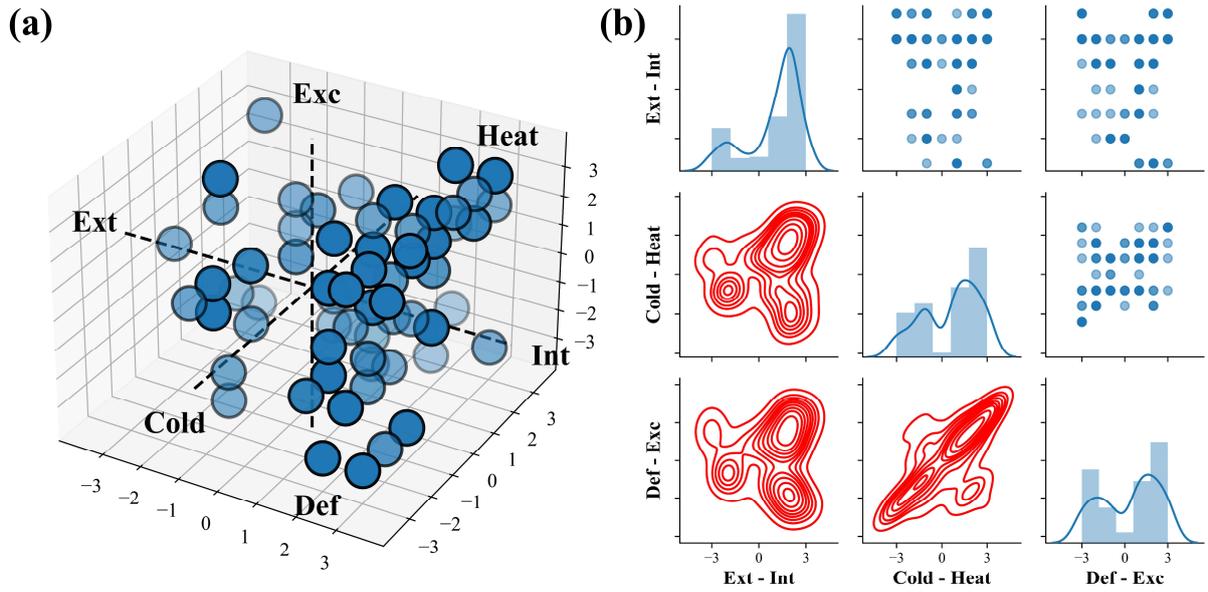

**Figure 1. The manifold of six principle patterns in *Shang-Han-Lun*.** (A) Each data point corresponds to a single provision from *Shang-Han-Lun*, with coordinate values in the 3D scatter plot representing the scores of each provision across the three dimensions (Exterior-Interior, Cold-Heat, Deficiency-Excess) based on the described symptoms. The resulting data manifold shows skewness depending on the subspaces. (B) The pair grid plot illustrates pairwise relationships between the three dimensions, with the diagonal showing 1D histograms, the lower triangular section displaying 2D kernel density plots, and the upper triangular section presenting 2D scatter plots.

**Table 1**

Variance for each dimension in the reduced pattern space

| Dimension | Variance |
|:---:|:---:|
| Ext – Int | 2.86 |
| Cold – Heat | 3.45 |
| Def – Exc | **4.05** |

2.1.2. Hypothesis II: The Exterior-Interior dimension represents the most abstract information about the patient's symptoms.

Given that the patterns in TEAM draw on metaphors, PI can be seen as a process of abstraction(9, 10). For highly abstract concepts, it is feasible to dichotomize an object and generalize over unseen instances(11). Therefore, we hypothesize that the abstract representation of the Ext-Int pattern contributes to the efficiency of the PI process and may explain the sequence used in PI.

To measure the extent of abstraction in each pattern, we quantified the degree of clustering using the abstraction index, which is the ratio between the average inter-group distance and the average intra-group distance(11). When data points are classified based on a binary label along a single dimension, or axis, an abstraction index greater than one suggests that the representations exhibit a specific geometric structure such as distinct clustering along that same axis. In our study, we constructed symptom vectors for each provision in *Shang-Han-Lun*, where each element indicates the presence or absence of a specific symptom by 1 or 0. The abstraction index calculated from these vectors was statistically significant only for the Ext-Int pattern, suggesting that symptom vectors form distinguishable clusters along the Ext-Int axis (**Table 2**).

**Table 2**

The abstraction index of each dimension in the symptom space

| Dimension | Abstraction Index |
|---|---|
| Ext – Int | **1.023\*** |
| Cold – Heat | 1.011 |
| Def – Exc | 1.011 |

Asterisk (\*) denotes statistical significance (p < 0.001).

While abstraction by clustering is one of the simplest indices of abstract representation, cross-conditional generalization performance (CCGP) serves as a more comprehensive index for evaluating generalization capacity, a key feature of abstraction(11). CCGP is analogous to cross-validated decoding performance, with the key difference being that the training and test data sets are distinctly conditioned. This allows us to evaluate whether the discrimination of a feature can be generalized even when conditions change. For example, to assess the generalization capacity for the Ext-Int classification, performance on a Heat-conditioned group (Ext-Heat, Int-Heat) would be measured using a classifier trained on a Cold-conditioned group (Ext-Cold, Int-Cold) (**Fig. 2**).

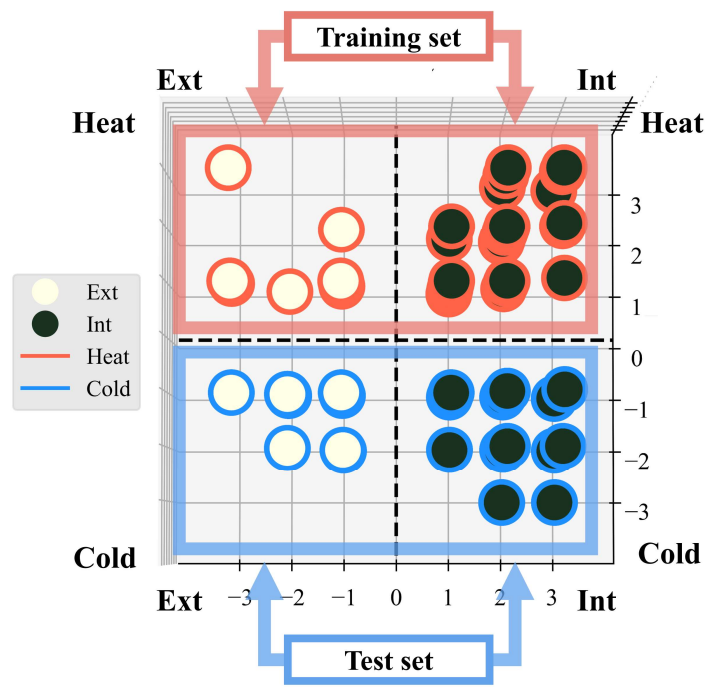

**Figure 2. Schematic example of measuring cross-conditional generalization performance to evaluate the generalization capacity.** Each data point belongs to one of four groups based on the Exterior-Interior (Ext-Int) and Cold-Heat pattern status. The classifier is trained with two groups sharing the Cold pattern status, labeled either Ext or Int. Its accuracy in classifying the Ext or Int label is then evaluated using test-set data from the other two groups sharing the Heat pattern status. Strong test-set performance suggests that representations in the Ext-Int dimension can be generalized across the Cold-Heat condition.

Consistent with the abstraction index, the generalization capacity for the Ext-Int dimension was higher than that for the other dimensions. Given that chance-level performance is 50%, the lower CCGP values for the Def-Exc dimension (53-56%) and the Cold-Heat dimension (48-52%) suggest that the model's ability to generalize in these dimensions is limited (**Table 3**). These findings support our second hypothesis, as the Ext-Int dimension demonstrates the highest abstraction index and generalization capacity, indicating that it represents the most abstract information about the patient's symptoms.

**Table 3**

Average CCGP of each dimension in the symptom space

| Training set | Test set | Label for classification | CCGP (%) |
| --- | --- | --- | --- |
| Ext/Heat, Int/Heat | Ext/Cold, Int/Cold | Ext - Int | **62** |
| Ext/Def, Int/Def | Ext/Exc, Int/Exc | Ext - Int | **60** |
| Ext/Cold, Int/Cold | Ext/Heat, Int/Heat | Ext - Int | **59** |
| Ext/Exc, Int/Exc | Ext/Def, Int/Def | Ext - Int | **58** |
| Def/Heat, Exc/Heat | Def/Cold, Exc/Cold | Def - Exc | 56 |
| Def/Cold, Exc/Cold | Def/Heat, Exc/Heat | Def - Exc | 54 |
| Ext/Def, Ext/Exc | Int/Def, Int/Exc | Def - Exc | 53 |
| Int/Def, Int/Exc | Ext/Def, Ext/Exc | Def - Exc | 53 |
| Ext/Cold, Ext/Heat | Int/Cold, Int/Heat | Cold - Heat | 52 |
| Int/Cold, Int/Heat | Ext/Cold, Ext/Heat | Cold - Heat | 51 |
| Def/Cold, Def/Heat | Exc/Cold, Exc/Heat | Cold - Heat | 51 |
| Exc/Cold, Exc/Heat | Def/Cold, Def/Heat | Cold - Heat | 48 |

All subgroups consist of 14 randomly selected samples, corresponding to the number of provisions in the smallest subgroup. The label represents the pattern that the classifier was trained to distinguish in both the training and test sets. CCGP is the prediction accuracy of the linear support vector machine measured on the test set. The averages of CCGPs obtained after 100,000 random samplings are presented in descending order.

2.1.3. Hypothesis III. The Exterior-Interior dimension facilitates the classification within the herbal space, aiding in the selection of appropriate prescriptions.

In the verification of the second hypothesis, we confirmed that the Ext-Int pattern is the most abstractly represented in the symptom space. Given that the fundamental purpose of PI is to select a prescription rather than merely classify patients, we established the third hypothesis to investigate the abstract representation of the three patterns in the herbal space. In this space, each patient is represented as a vector of 170 dimensions, corresponding to the number of herbs mentioned in *Shang-Han-Lun*, with each element indicating the presence of the corresponding herb. We focused on the possibility that the latent pattern space emerged not only to compress the symptom space but also to efficiently map symptom space to herbal space.

To intuitively explore whether information for herb selection is represented in the pattern space, the data points in the pattern space were color-coded based on the similarity of prescriptions at the herb level (**Fig. 3**). Consistent with previous analyses, the groups were visually distinguishable along the Ext-Int dimension. The Ext group appeared relatively homogeneous, while the Int group showed more heterogeneity. These results suggest that herb-related information is meaningfully represented in the pattern space.

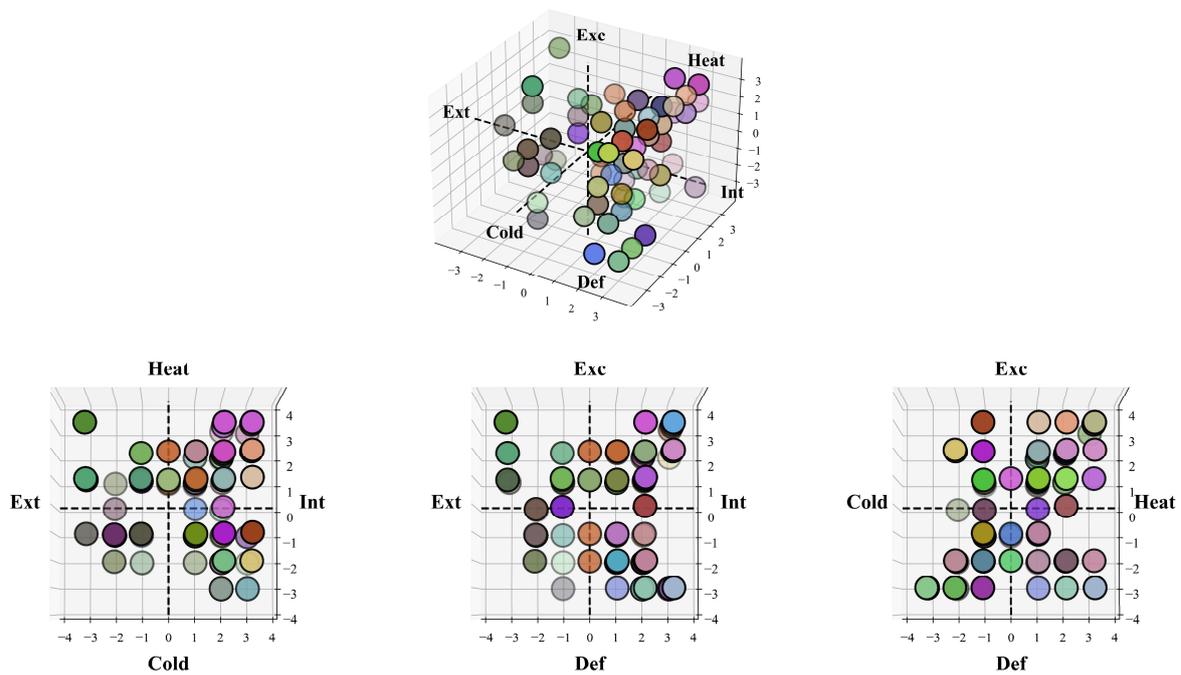

**Figure 3**. **Color-coded pattern manifolds of provisions in *Shang-Han-Lun* representing the prescription similarity.** Different cross-sections of the manifolds were visualized for comparison.

The degrees of abstraction of three patterns in the herbal space were analyzed using the abstraction index. All three patterns were abstractly represented with statistical significance (**Table 4**). CCGP analysis revealed that the Ext-Int pattern could be generalized the most in the herbal space, as with the symptom space (**Table 5**).

**Table. 4**

The abstraction index of each dimension in the herbal space

| Dimension | Abstraction Index |
|---|---|
| Ext-Int | **1.085*** |
| Cold-Heat | 1.044* |
| Def-Exc | 1.026* |

Asterisk (*) denotes statistical significance (p < 0.001).

**Table 5**

Average CCGP of each dimension in the herbal space

| Training set | Test set | Label for classification | CCGP (%) |
|---|---|---|---|
| Ext/Exc, Int/Exc | Ext/Def, Int/Def | Ext - Int | **72** |
| Ext/Heat, Int/Heat | Ext/Cold, Int/Cold | Ext - Int | **66** |
| Ext/Def, Int/Def | Ext/Exc, Int/Exc | Ext - Int | **65** |
| Ext/Cold, Int/Cold | Ext/Heat, Int/Heat | Ext - Int | **57** |
| Def/Cold, Exc/Cold | Def/Heat, Exc/Heat | Def - Exc | 56 |
| Def/Heat, Exc/Heat | Def/Cold, Exc/Cold | Def - Exc | 53 |
| Int/Def, Int/Exc | Ext/Def, Ext/Exc | Def - Exc | 53 |
| Ext/Def, Ext/Exc | Int/Def, Int/Exc | Def - Exc | 52 |
| Def/Cold, Def/Heat | Exc/Cold, Exc/Heat | Cold - Heat | 51 |
| Exc/Cold, Exc/Heat | Def/Cold, Def/Heat | Cold - Heat | 51 |
| Ext/Cold, Ext/Heat | Int/Cold, Int/Heat | Cold - Heat | 47 |
| Int/Cold, Int/Heat | Ext/Cold, Ext/Heat | Cold - Heat | 45 |

All subgroups consist of 14 randomly selected samples, corresponding to the number of provisions in the smallest subgroup. The label represents the pattern that the classifier was trained to distinguish in both the training and test sets. CCGP is the prediction accuracy of the linear support vector machine measured on the test set. The averages of CCGPs obtained after 100,000 random samplings are presented in descending order.

2.2. Validation

2.2.1 Decision tree regression

Having demonstrated that the Ext-Int pattern shows the highest abstraction and generalization capacity in both the symptom and herbal spaces, we sought to validate whether this competence translates into practical utility in the clinical decision-making process. To this end, we developed a decision tree model to regress herb vectors onto symptom vectors and assess the utility of Ext-Int dimension in selecting appropriate prescriptions.

The decision tree is an algorithm for multistage decision-making, decomposing a complex decision into a series of simpler decisions(12). It also represents the underlying rules in data with a sequential structure that recursively splits the data, which is analogous to the process of PI. Features are selected in a sequential order based on data exploration, taking into account information gain and computational efficiency in classification(13). By examining the top nodes selected in the tree, we aimed to identify which features of the input vectors were most influential in predicting prescriptions and what patterns they imply. The decision tree was trained to estimate 170-dimensional herb vectors using 702-dimensional symptom vectors. For each given tree depth, the top three symptom nodes are presented along with their feature importance (**Table 6**). The selected symptoms are detailed in **Supplementary Table 1**, represented by Chinese characters and descriptions based on the WHO International Standard Terminologies.

Floating pulse (浮脈) was repeatedly selected as the root node at several depths. In TEAM, floating pulse is a typical symptom of the Ext pattern, indicating that external pathogens have not yet penetrated deeply into the body(7). Spontaneous urination (小便自利), often selected as the second node, also reflects the depth of the pathogenic factor (i.e., Ext-Int). In *Shang-Han-Lun*, spontaneous urination is a contextual cue suggesting that the internal functions of the body remain intact. Together, these data-driven selections of symptoms related to the Ext-Int pattern support the idea that primary discrimination on the Ext-Int pattern contributes to the effective mapping between symptoms and herbs.

**Table 6**

Results of decision tree regression: 702-dimensional symptom vector for input

| Depth | 1st Selected Features | 2nd Selected Features | 3rd Selected features | No. of features | $R^2$ |
|---|---|---|---|---|---|

| Depth | 1st Selected Features | | 2nd Selected Features | | 3rd Selected features | | No. of features | $R^2$ |
|---|---|---|---|---|---|---|---|---|
| 3 | Floating pulse | 0.23 | Spontaneous urination | 0.21 | Aversion to heat | 0.16 | 6 | 0.08 |
| 5 | Floating pulse | 0.12 | Spontaneous urination | 0.10 | Aversion to heat | 0.08 | 18 | 0.17 |
| 7 | Floating pulse | 0.08 | Spontaneous urination | 0.07 | Dry retching | 0.06 | 27 | 0.24 |
| 10 | Floating pulse | 0.06 | Diarrhea | 0.05 | Spontaneous urination | 0.05 | 41 | 0.35 |
| 30 | Floating pulse | 0.03 | Spontaneous urination | 0.03 | Alternating chills and fever | 0.02 | 107 | 0.77 |

Number (No.) of features and the coefficient of determination ($R^2$) are derived from the in-sample performance.

Additionally, when herb vectors were regressed using the symptom vectors concatenated with the corresponding 3-dimensional pattern vectors, the Ext-Int and Cold-Heat pattern features were consistently chosen for the root and second nodes, respectively (**Table 7**). The feature importance of the root node increased substantially at each depth compared to the results obtained using symptom-only vectors (i.e., without incorporating the pattern features), and in-sample performance also improved. This suggests that patterns are more informative for prescription decisions than individual symptoms alone, indicating that PI can improve the accuracy of prescription choices. In conclusion, the results of decision tree regression support the notion that the sequential structure of PI, as well as PI itself, facilitates the choice of appropriate prescriptions.

**Table 7**

Results of decision tree regression: 702-dimensional symptom vector concatenated with 3-dimensional pattern vector for input

| Depth | 1st Selected Features | | 2nd Selected Features | | 3rd Selected features | | No. of features | $R^2$ |
|---|---|---|---|---|---|---|---|---|
| 3 | Ext-Int | 0.57 | Cold-Heat | 0.23 | Spontaneous urination | 0.07 | 5 | 0.16 |
| 5 | Ext-Int | 0.31 | Cold-Heat | 0.13 | Dry retching | 0.06 | 20 | 0.29 |
| 7 | Ext-Int | 0.24 | Cold-Heat | 0.10 | Dry retching | 0.04 | 37 | 0.38 |
| 10 | Ext-Int | 0.19 | Cold-Heat | 0.08 | Dry retching | 0.03 | 56 | 0.48 |
| 30 | Ext-Int | 0.10 | Cold-Heat | 0.04 | Fever | 0.03 | 119 | 0.88 |

Number (No.) of features and the coefficient of determination ($R^2$) are derived from the in-sample performance.

## 3. Discussion

This study demonstrates that the clinical decision-making processes in TEAM can be effectively modeled and validated using quantitative techniques, particularly through the lens of dimensionality reduction. Our results empirically confirm that the Ext-Int dimension plays a crucial role in the PI process, showing the highest abstraction and generalization capacity in both the symptom and herbal spaces. This highlights the value of differentiating highly abstract features early in diagnosis. In the CCGP analysis, inferences about the Ext-Int pattern were less affected by conditions such as Cold-Heat or Def-Exc pattern. This suggests that it remains easier to distinguish the Ext-Int pattern in clinical practice even when these mixed conditions are present, allowing

TEAM doctors to make more robust diagnostic inferences. Moreover, our analyses in the herbal space revealed that, while the Ext-Int pattern was the most abstract and generalizable, the Cold-Heat and Def-Exc patterns also contributed their valuable insights, showing statistically significant abstraction values and especially Cold-Heat serving as a key feature in the decision tree regression. These results underscore the layered utility of all three patterns when mapping symptoms to herbal prescriptions.

From the perspective of TEAM practice, this outcome may be attributed to the clinical and conceptual clarity and simplicity of the Ext-Int pattern in comparison to the other two patterns under consideration. For example, in the WHO International Standard Terminologies for Traditional Chinese Medicine, which functions as a comprehensive dictionary for TEAM terminology, the number of terms including the six patterns in EPPI across three key domains—theoretical fundamentals, diagnosis and patterns, and treatment principles and methods—is the lowest for the Ext-Int pattern, followed by the Def-Exc pattern, and then the Cold-Heat pattern(7). This distribution suggests that the Exterior-Interior pattern is clinically and conceptually more straightforward, whereas the Cold-Heat and Deficiency-Excess patterns exhibit greater complexity. The differences in the characteristics of these patterns within the TEAM system could be further explored in future research.

This study has limitations that should be considered. Our analyses were conducted using only clinical cases from *Shang-Han-Lun*, which may hinder the generalization of insights into TEAM doctors' decision-making processes across broader clinical settings. Nonetheless, *Shang-Han-Lun* remains one of the most representative clinical texts in TEAM, with its influence and applicability being substantial even at present(14, 15). Future studies using more diverse TEAM datasets are expected to further validate and support the findings of this research.

The ultimate goal of this study is to model what it means to think like a TEAM doctor and to gain insights from that model. By bridging fields such as machine learning, computational neuroscience, and cognitive neuroscience, we will be able to develop new operational definitions and powerful tools for quantitative analyses of the clinical decision-making process in TEAM doctors. These tools will form the basis for quantified and standardized TEAM processes, contributing to advancements in AI-driven services both in TEAM and conventional medicine. Additionally, incorporating explainable AI into this framework can create a positive feedback loop: explainable AI can enhance the reliability of medical AI systems, while these systems, in turn, will promote advancements in medicine, including TEAM. This multidisciplinary approach not only deepens our understanding of the cognitive processes of TEAM doctors but also offers fresh insights into the essence of TEAM itself. Ultimately, this study presents a novel approach to exploring TEAM and has the potential to contribute to the advancement of clinical practice, education, and research in the field of medicine.

## 4. Methods

4.1. Pattern manifold in *Shang-Han-Lun*

A total of 242 provisions which include prescriptions were extracted from *Shang-Han-Lun*. For each provision, four TEAM doctors independently evaluated scores across three dimensions (Ext-Int, Cold-Heat, Def-Exc) using a 7-point scale. In cases of disagreement, a consensus was reached through discussion. The reduced pattern space was expressed as a 3-dimensional space, with each axis representing one of the Ext-Int, Cold-Heat, and Def-Exc patterns, and each provision represented as a data point.

4.2. Abstraction index

The abstraction index was used to quantify the degree of clustering in each pattern, defined as the ratio between the average inter-group distance and the average intra-group distance:

$$\text{Abstraction index} = \frac{\frac{1}{n_x n_y}\sum_{(x,y)} dist(\boldsymbol{x}, \boldsymbol{y})}{\alpha_x \sum_x dist(\boldsymbol{x}, \bar{x}) + \alpha_y \sum_y dist(\boldsymbol{y}, \bar{y})}$$

Where $\boldsymbol{x}$ and $\boldsymbol{y}$ represent vector data, $n_x$ and $n_y$ are the number of data points in the groups to which $\boldsymbol{x}$ and

$y$ belong, $dist(x, y)$ is the Euclidean distance between $x$ and $y$, and $\alpha_x$ and $\alpha_y$ are the proportions of data in the $x$ and $y$ groups relative to the total data.

Symptom vectors were constructed similarly to herb vectors, except that symptoms of the same category were grouped together (e.g., aversion to cold [惡寒] and fear of cold [畏寒] were treated as the same symptom). The significance of the obtained values was verified using permutation tests, where binary labels for each dimension of the provisions were randomly shuffled 10,000 times to calculate p-values.

4.3. CCGP

To evaluate the generalization capacity of the identified patterns, a linear support vector machine was trained to discriminate labels for one of the three patterns (Ext-Int, Cold-Heat, Def-Exc). Classification accuracy was measured on a test set, which shared a disjoint condition against the training set (i.e., unseen during training). Due to the disparity in the number of data across the 12 subgroups formed by combinations of the three patterns, 14 instances were randomly sampled from each group—the number of data in the smallest subgroup—to correct for label imbalance. The average performance was reported after 100,000 iterations, accounting for randomness introduced during sampling.

4.4. Color coding for indicating similarity between prescriptions

To explore whether herb information was represented in the pattern space, data points were color-coded according to the prescription similarity of each provision (as shown in Fig. 3). For this purpose, a 170-dimensional herb vector was constructed for each provision. Each herb mentioned in *Shang-Han-Lun* was encoded as a binary variable (0 = not included, 1 = included). These herb vectors were then compressed into three dimensions, preserving the distance relationships between them through multi-dimensional scaling(16). After normalization, the values in the obtained 3-dimensional vectors were inputted into the red, green, and blue channels to color each data point, such that closely located prescriptions in the herbal space were represented by similar colors on the pattern manifold.

4.5. Decision tree regression

The decision tree was trained to regress 170-dimensional herb vectors using the 702-dimensional symptom vectors as input. The branching criterion was a mean squared error. Results were reported across a range of branching depths (3-30), including the top three features selected at each depth, their feature importance, the number of features used, and the coefficient of determination ($R^2$) as a measure of in-sample performance. It is important to note that the symptom vectors used here were 702-dimensional, comprising all symptoms mentioned in *Shang-Han-Lun*, without categorization.

All analyses were performed using Python and the scikit-learn library(17).